\pdfoutput=1

\documentclass[11pt]{article}

\usepackage[]{ACL2023}

\usepackage{times}
\usepackage{latexsym}

\usepackage[T1]{fontenc}

\usepackage[utf8]{inputenc}

\usepackage{microtype}
\usepackage{inconsolata}
\usepackage{amsbsy}

\newcommand\myeq{\mkern1.5mu{=}\mkern1.5mu}

\usepackage{xcolor}

\newlength{\starlen}
\usepackage{amsmath}
\usepackage{amsthm}
\usepackage{amsfonts}
\usepackage{amssymb}
\usepackage{mathtools}

\newcommand{\sigmoid}{\mathrm{sigmoid}}

\newcommand{\vtheta}{{\boldsymbol \theta}}
\newcommand{\vphi}{{\boldsymbol \phi}}

\mathchardef\mhyphen="2D

\newcommand{\calD}{{\cal D}}

\DeclarePairedDelimiterX{\infdivx}[2]{(}{)}{%
  #1\;\delimsize\|\;#2%
}

\newcommand{\expnumber}[2]{{#1}\mathrm{e}{#2}}

\usepackage{todonotes}
\makeatletter
\newcommand*\iftodonotes{\if@todonotes@disabled\expandafter\@secondoftwo\else\expandafter\@firstoftwo\fi}
\makeatother

\definecolor{edolime}{rgb}{0.9,1,0.3}

\definecolor{lightblue}{rgb}{0.7,0.85,1}

\definecolor{orange}{rgb}{1,0.6,0.2}

\usepackage{tikz}
\usetikzlibrary{bayesnet}
\usepackage{graphicx}
\usepackage{booktabs}
\usepackage{tabularx}
\usepackage{tabulary}
\usepackage{colortbl}
\usepackage{xspace}
\usepackage{xcolor}
\newcolumntype{Y}{>{\centering\arraybackslash}X}
\usepackage{multirow}
\usepackage{algorithm}
\usepackage[noend]{algorithmic}
\usepackage[capitalize,noabbrev]{cleveref}
\title{Efficient Transformers with Dynamic Token Pooling}

\author{Piotr Nawrot$^\dagger$~~~~~Jan Chorowski$^\ddagger$~~~~~Adrian Łańcucki$^{\diamond\clubsuit}$~~~~~Edoardo M. Ponti$^\dagger$ \\
$^\dagger$University of Edinburgh~~~~~$^\ddagger$Pathway~~~~~$^\diamond$NVIDIA~~~~~$^\clubsuit$University of Wrocław\\
\texttt{piotr.nawrot@ed.ac.uk}
}

\begin{document}
\maketitle

\begin{abstract}
Transformers achieve unrivalled performance in modelling language, but remain inefficient in terms of memory and time complexity. A possible remedy is to reduce the sequence length in the intermediate layers by pooling fixed-length segments of tokens. Nevertheless, natural units of meaning, such as words or phrases, display varying sizes. To address this mismatch, we equip language models with a dynamic-pooling mechanism, which predicts segment boundaries in an autoregressive fashion. We compare several methods to infer boundaries, including end-to-end learning through stochastic re-parameterisation, supervised learning (based on segmentations from subword tokenizers or spikes in conditional entropy), as well as linguistically motivated boundaries. We perform character-level evaluation on texts from multiple datasets and morphologically diverse languages. The results demonstrate that dynamic pooling, which jointly segments and models language, is both faster and more accurate than vanilla Transformers and fixed-length pooling within the same computational budget.
\end{abstract}

\section{Introduction}
The Transformer architecture \citep{vaswani2017attention} lies at the heart of cutting-edge generative models, such as GPT-3 \citep{brown2020language} for text and DALL·E 2 \citep{ramesh2022hierarchical} for images. Its success can be largely attributed to the ability to leverage a considerable amount of data, which yields performance gains \citep{kaplan2020scaling} and emergent abilities \citep{wei2022emergent} in accordance with well-established scaling laws. Nonetheless, the time and memory efficiency of Transformers remains constrained by their algorithmic complexity of $\mathcal{O}(l^2n)$, where $l$ stands for sequence length and $n$ for the number of layers.

To remedy this shortcoming without renouncing the expressivity of a deep model, the quadratic self-attention can be sparsified \citep{child2019generating,roy-etal-2021-efficient,ren2021Combiner} or linearly approximated \citep{Beltagy2020Longformer}. 
Hourglass Transformers \cite{nawrot2021hierarchical} provide an alternative solution, where the sequence length is reduced in the intermediate layers by merging fixed-size groups of tokens, similar to \citep{dai2020funnel}. These pooled representations are up-sampled back to the original length in order to generate sequences in an auto-regressive fashion \citep{ronneberger2015u}.

Nevertheless, pooling groups of fixed size is sub-optimal in several respects. First, these groups are misaligned with linguistic primitives: units of meaning such as morphemes, words, phrases, and clauses vary in size. Second, the elements of a sequence may carry different degrees of information (for instance, silence and voice in speech). Ideally, the model should perform \textit{hierarchical} computation, relying on the same abstractions as human processing of language, and \textit{conditional}, by allocating resources to sub-sequences in proportion to the model uncertainty.
In this work, we demonstrate that dynamic pooling results not only in higher shortening rates of input sequences, and thus increased efficiency, but also superior performance in next token prediction due to adopting the correct inductive bias in grouping tokens.

 \begin{figure*}[tb]
     \centering
     \includegraphics[width=0.85\linewidth]{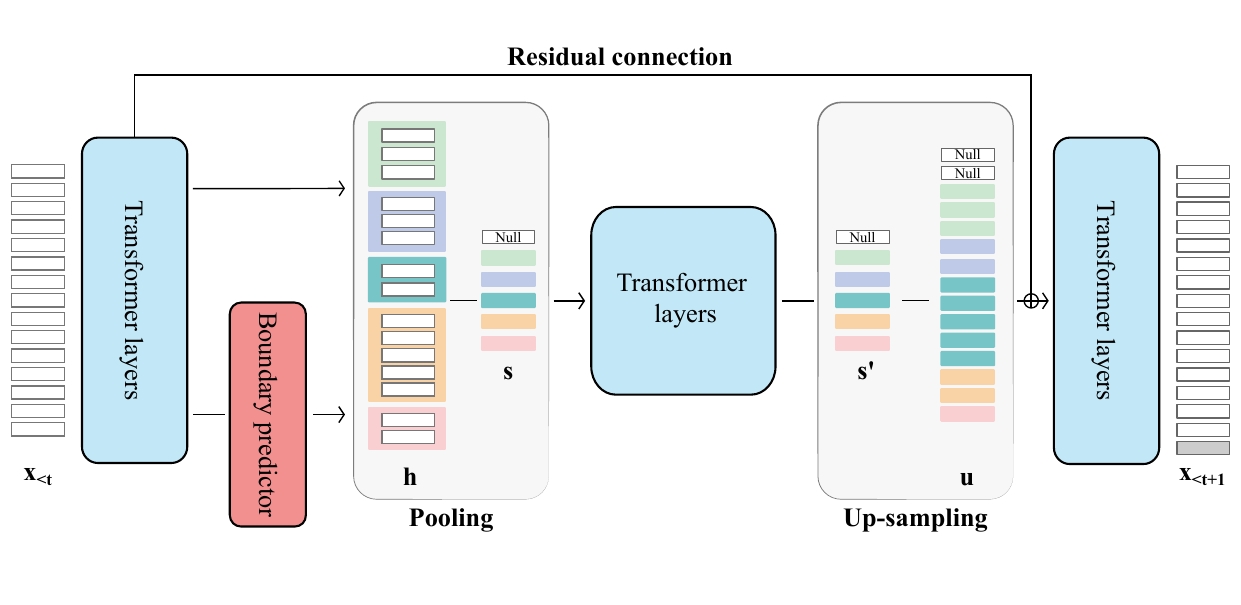}
     \vspace{-4mm}
     \caption{The architecture of a dynamic-pooling Transformer, which jointly performs language modelling and token segmentation. The boundary predictor predicts segment boundaries and pools together groups of variable length by averaging. The shortened sequence is processed efficiently by a series of intermediate layers, then up-sampled back to the original length via duplication. The model generates the next token $\pmb{x}_t$ in the same resolution as the input.}
     \label{fig:model}
 \end{figure*}

To this end, we propose a new Transformer variant that jointly learns token sequences and dynamically pools them into latent groupings of variable size (\cref{fig:model}). Crucially, the segmentation must preserve the auto-regressive property, and typical subword tokenizers cannot be applied to incomplete sequences during generation. Rather, we learn a neural boundary predictor during training: 1) supervised by tokenizers such as Unigram \citep{kudo2018subword}; 2) supervised by spikes in the conditional entropy of the predictive distribution, which ensure that the computation is adaptive to the level of uncertainty of the sequence model; 3) end-to-end through stochastic re-parameterisation \citep{maddison2017the,jang2017categorical}; 4) use natural data boundaries such as whitespaces, which separate words in many scripts, without a predictor.

To validate our model, we experiment with character-level language modelling of text in several English benchmarks, including \texttt{text8} \citep{text8}, \texttt{CC-100} \citep{wenzek-etal-2020-ccnet}, and \texttt{wiki40b} \citep{guo-etal-2020-wiki}, as well as in a series of languages representing different morphological types: Finnish, Hebrew, and Vietnamese. We find that dynamic pooling not only achieves lower time and memory complexity, but even surpasses the performance of vanilla Transformers and fixed-size pooling Transformers in most benchmarks by statistically significant margins.

Overall, our results indicate a promising direction to further accelerate training and therefore facilitate scaling. A FAQ section about our methods, findings, and the experimental setup is available in Appendix~\ref{app:faq}. We release the code at \url{https://github.com/PiotrNawrot/dynamic-pooling}.

\section{Background}

\subsection{Language Modelling with Transformers}
Let $\pmb{x}=(x_1, \ldots, x_l)$ denote the input sequence.
A language model assigns a probability value to any possible sequence of tokens from a vocabulary $\mathcal{V}$.
The parameters of a model $\vtheta$ are optimised to maximise the aggregate probability of all $\pmb{x} \in \mathcal{V}^*$ in the training set $\calD$:
\begin{equation} \label{eq:nextcharprob}
\mathop{\mathrm{arg max}}_\vtheta \sum_{\pmb{x} \in \calD} \sum_{t=1}^l \log p(x_t \mid \pmb{x}_{< t}, \vtheta),
\end{equation}
where \textit{t} indexes time steps. In our experiments, $\vtheta$ corresponds to the parameters of an autoregressive Transformer model \citep{vaswani2017attention}.

A key advantage of Transformers is their ability to scale, which ultimately reaps the largest benefits according to \citep{sutton2019bitter}'s `bitter lesson' and reveals surprising emergent capabilities of language models \citep{kaplan2020scaling,wei2022emergent}.
Nevertheless, the algorithmic complexity of self-attention, $\mathcal{O}(l^2)$ where $l$ is the length of the sequence, creates a bottleneck. To alleviate this cost, previous work \citep{clark2021canine,tay2021charformer,nawrot2021hierarchical} proposed to reduce the sequence length after the initial layers by pooling together groups of tokens. A single shortening by a factor $k$ reduces the complexity to $\mathcal{O}(\frac{l^2}{k^2})$. This allows for increasing either the model efficiency or its depth within the same compute budget.

\subsection{Hourglass Transformer}
\label{sec:hourglass}
Na\"ive length reduction through pooling would reduce the length of output, however language models operate with the same input and output resolutions. For this reason, \citep{nawrot2021hierarchical} introduced the Hourglass Transformer composed of three blocks of Transformer layers, which downsample, process, and upsample the tokens back to the original granularity.
The first block encodes each input token ${x}_t$ into $\pmb{h}_t$. Afterwards, groups of adjacent tokens of fixed length $k$ are mean-pooled to form $\lceil \frac{l}{k} \rceil$ representations ${\pmb{s}}$:
\begin{equation}
    {\pmb{s}}_m = \frac{1}{k} \sum_{i=mk-k+1}^{mk} \pmb{h}_i
\end{equation}
Next, each pooled representation ${\pmb{s}}_m$ is processed by the middle block of Transformer layers, which operates with complexity $\mathcal{O}(\frac{l^2}{k^2})$, yielding $\pmb{s}_m^\prime$. This sequence is up-sampled to its original resolution by duplication: ${{\pmb{u}}}_t = {\pmb{s}^\prime}_{\lceil \frac{t-k+1}{k} \rceil}$, and added to the hidden representations $\pmb{h}$ from before shortening through a skip connection, and passed to the third block.

Note that we subtract $k-1$ from the index. This is because pooling and up-sampling in an autoregressive model pose a risk of data leakage from the future to the past. In fact, up-sampled representations might encompass future tokens if no measures are taken to prevent this. As a remedy, Hourglass Transformer shifts the up-sampled sequence to the right by $k-1$ positions, and pads it with a learnable null-group representation ${{\pmb{u}}}_0$ at the beginning. This is sufficient to satisfy the autoregressive property in the fixed pooling scenario.\footnote{We refer to  \citep{nawrot2021hierarchical} for more details.}

Hourglass Transformer was shown to improve time and space complexity in a number of language and image modelling tasks, for a given parameter count. However, this came at the expense of degrading the perplexity of the language model, especially with shortening factors $k > 2$. We conjecture that this undesirable side effect is due to two main reasons. Firstly, the distribution of lengths of natural units of meaning such as morphemes and phrases in natural languages is uneven: for instance, word length is correlated with its frequency \citep{zipf2016human,bentz2016zipf}. Secondly, information content tends to be distributed uniformly across units of meaning \citep{meister-etal-2021-revisiting}.

As a consequence, fixed pooling creates segments with incongruous boundaries and unequal information content. For instance, in speech, this results in giving silence and voice the same importance. Instead, an ideal model should allocate compute \textit{conditionally} on the information content of a given token. This would also ultimately lead to interpreting language \textit{hierarchically} based on the same abstractions that humans adopt for language processing. Hence, we present a method to enable variable-length pooling and up-sampling in autoregressive language models.

\section{Dynamic-Pooling Transformer}

\subsection{Boundary Prediction}
In order to augment the Hourglass architecture with variable-size pooling, we seek to find a sequence of segment boundaries $\pmb{b}\in\{0,1\}^l$ for every input $\pmb{x}$. Let $b_t = 1$ denote a segment boundary between elements $x_t$ and $x_{t+1}$. The boundary predictor is implemented as a Multi-Layer Perceptron with parameters $\vphi$. As shown in \cref{fig:model}, this module maps each representation $\pmb{h}_t$ encoded by the first stack of Transformer layers into a Bernoulli probability distribution:
\begin{equation} \label{eq:boundaryprob}
    \hat{b}_t = p(b_t\myeq 1) = \sigmoid\left( \mathrm{MLP}_\vphi\left(\pmb{h}_t\right) \right).
\end{equation}
Since segment boundaries are discrete, sampling from this distribution is not differentiable with respect to the model perplexity. Hence, we optimise this latent variable through stochastic re-parametrisation \citep{jang2017categorical,maddison2017the} via hard Gumbel-sigmoid (\cref{sssec:gs}), jointly learning the language model and boundary predictor.
We favour this solution over a score-function estimator of the gradient, as it suffers from high variance and computation costs due to sampling \citep{schulman2015gradient}. 

As an alternative, we explore training the boundary predictor module with a binary cross-entropy loss with respect to two different sources of supervision: a Unigram tokenizer (\cref{sssec:tok}) and spikes in conditional entropy (\cref{sssec:es}). Finally, we consider resorting to linguistically inspired boundaries (\cref{sssec:ws}). 
During training and evaluation, we perform maximum likelihood inference for these variables. In other words, each $\hat{b}_t$ from \cref{eq:boundaryprob} is rounded to the closest binary scalar such that $b_t = \lfloor \hat{b}_t \rceil$.

\begin{figure}[tb]
    \centering
    \includegraphics[scale=1.15]{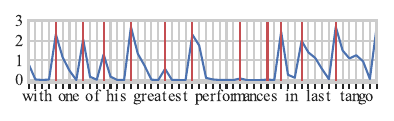}\\
    \includegraphics[scale=1.15]{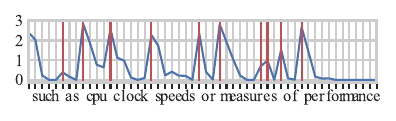}
    \caption{Entropy of a Transformer character-level language model in two text segments. Red vertical lines indicate the boundaries according to spikes in conditional entropy. Most of them coincide with whitespaces, due to the high uncertainty at word starts, but they also fall after morphemes like `\textit{great}' or `\textit{measure}'. Segmentation may vary based on the context, e.g., of the word `\textit{performance}'.}
    \label{fig:entropy}
\end{figure}

\subsubsection{Segmenting with Gumbel-Sigmoid}
\label{sssec:gs}
In order to learn the input segmentation end-to-end based on the model perplexity, we can re-parameterise the Bernoulli distribution of \cref{eq:boundaryprob} by injecting stochasticity in this form:
\begin{align} \label{eq:allocation}
    \hat{b}_{t} &=  \textrm{sigmoid} \left[ \log \frac{\hat{b}_t \, u}{(1 - \hat{b}_t) \, (1 - u)}^{1/\tau}\right] \nonumber \\ u &\sim \mathrm{Uniform}(0, 1).
\end{align}
where $\tau$ is the temperature, a hyper-parameter. This estimator, however, is biased and might lead to sub-optimal results. As a consequence, we also propose methods based on supervised learning of the boundary predictor in the following sections.

\subsubsection{Segmenting with Subword Tokenizers}
\label{sssec:tok}
Widespread algorithms for extracting variable-length boundaries for text are subword tokenizers, including Unigram \cite{kudo2018subword}, Byte Pair Encoding ~\cite[BPE;][]{sennrich2016neural}, and WordPiece~\cite{schuster2012japanese}. However, these create subwords greedily, and might change the segmentation of a given sequence prefix after more tokens are observed. For instance, consider the phrase `\textit{civil aviation}'. A Unigram model might segment its prefix `\textit{civil aviatio}' differently before and after observing the character `\textit{n}':\\

\begin{tabular}{l}
\hspace{3em}\texttt{\_civil \_a vi \textbf{ati o}}\\
\hspace{3em}\texttt{\_civil \_a vi \textbf{ation}}
\end{tabular}\\

\noindent
During training an entire sentence is tokenized, but during inference a prefix is extended one character at a time and re-tokenized, possibly changing the boundaries like in the example above. Hence, deploying off-the-shelf tokenizers na\"ively during inference does not recover the oracle segments and creates a mismatch between training and evaluation boundaries.

As a remedy, we provide the training tokenization as supervision to our autoregressive boundary predictor instead. More specifically, we employ a Unigram tokenizer~\cite{kudo2018subword}, as it aligns with morphological units better than other algorithms~\cite{bostrom2020byte}. To prevent subword units from crossing word boundaries, we split the text on whitespace characters beforehand. Vocabulary size is a tunable hyper-parameter which offers different efficiency--performance trade-offs.

\subsubsection{Segmenting with Entropy Spikes}
\label{sssec:es}
As an alternative to providing supervision through Unigram, we also propose a new segmentation method based on spikes of conditional entropy, which is agnostic about the presence of natural boundaries (such as whitespaces) or the availability of tokenizers. These properties make it suitable for other modalities in addition to text, such as speech and vision. Moreover, this enables top-down supervision and end-to-end training without external dependencies.

Intuitively, in natural language the information content tends to be spread evenly throughout a sentence, to facilitate communication. The conditional entropy is the expectation of such information content over the tokens in the vocabulary:
\begin{equation}
    \mathcal{H}(x_t \mid \pmb{x}_{<t}) = \sum_{x \in \mathcal{V}} p(x_t \mid \pmb{x}_{<t}) \underbrace{\left(- \log p(x_t \mid \pmb{x}_{<t})\right)}_{\text{information content}}
\end{equation}
Therefore, peaks in this conditional entropy provide indications of surprisal, and can serve as natural boundaries between segments. More formally, let $\mathcal{H}_t$ be the conditional entropy at time $t$. We select local spikes by comparing their value within a (left) window of size $k$. We place boundaries according to the following conditions:
\begin{equation}
 b_t = \begin{cases}
   1 & \text{ if }\ \mathcal{H}_t > \mathcal{H}_{i} \quad \forall i \in \{t-k, \dots, t-1\} \\
   0 & \text{ otherwise}.
\end{cases}
\end{equation}
Empirically, entropy spikes in language models overlap with word boundaries to a significant degree~\cite{hutchens1998finding}. However, they are also more flexible as they enable conditional computation based on the model's confidence about its next token prediction. As an example of segmentation based on entropy spikes, consider \cref{fig:entropy}.

\subsubsection{Linguistically Inspired Segments}
\label{sssec:ws}
Finally, perhaps the most straightforward source of segmentation is word boundaries. In fact, in many scripts, these are marked by whitespace characters.\footnote{Several scripts such as Chinese characters, however, do not adopt this convention.} The simplicity of this method of segmentation comes with the obvious drawback of not providing control over the rate of shortening, while we found that the optimal rate varies with the language. Hence its efficiency--performance trade-off is not tunable.

Segment boundaries are placed in between two symbols. In our experiments, we put a boundary \emph{after} a whitespace character. Thus, we do not need to train a boundary predictor, since predicting a whitespace character is a signal to close the group in the next iteration of auto-regressive generation. This would not be possible, had we chosen to put a boundary before a whitespace character.

\subsection{Pooling and Up-sampling}
In the pooling step (\cref{fig:model}) a generated sequence of boundaries $\pmb{b}$ is used to pool the tokens belonging to the same segment by averaging. Thus, we form $\sum_{t=1}^{l} b_t + 1$ shortened representations ${\pmb{s}}$, which are then passed to the middle block of Transformer layers. Note that for Gumbel-sigmoid, to keep pooling differentiable, we algebraically manipulate ${{\pmb{b}}} \in \mathbb{R}^{l}$ into $B \in \mathbb{R}^{l \times 1 + \sum_t b_t}$, i.e.\ a binary matrix that maps from the original length to the shortened length, following \citep{bhati2021segmental}. The cell $B_{ij}$ is 1 if token $i$ is merged into the $j$-th group, and 0 otherwise. Thus, $\pmb{s} = \pmb{h}B / \sum_i B_{i\star}$, where the denominator unit-normalises the matrix columns.

To obtain the up-sampled representation ${{\pmb{u}}}_t$ while preserving the autoregressive property, we calculate the largest index $m$ so that the output of the middle block ${\pmb{s}}^\prime_m$ does include future information: ${{\pmb{u}}}_t = {\pmb{s}}^\prime_m$, where $m = \sum_{i=1}^t b_i$.
As a consequence, a segment representation ${\pmb{s}}^\prime_m$ can only be added to the last token pooled into group $m$. For all the other non-final tokens, we take the representation of a previous segment ${\pmb{s}}^\prime_{m-1}$. Similar to Hourglass, the representation for the first (null) group ${{\pmb{s}}}_0$ is a learnable vector. Afterwards, ${{\pmb{u}}}_t$ is added to the highway layer representation $\pmb{h}_t$.

\subsection{Auxiliary Objectives}
In addition to minimising the language modelling loss with respect to the parameters $\vtheta$ as shown in \cref{eq:nextcharprob}, we use auxiliary objectives to train the boundary predictor parameters $\vphi$. For supervised learning with subword tokenizers and entropy spikes, we minimise the cross-entropy between predicted boundaries $\pmb{b}$ and gold ones. For end-to-end learning with Gumbel softmax, we introduce a regularizer based on a Binomial prior. Let $k = \sum_t b_t$:
\begin{equation}
    \mathrm{Binomial}(\alpha; l, k) = \binom{l}{k} \alpha^{k} (1 - \alpha)^{l - k}
\end{equation}
where $\alpha \in [0, 1]$ is a hyper-parameter. This regularizer prevents the model from collapsing into trivially predicting each position as a boundary.

\section{Experimental Setup}

\subsection{Datasets}
In addition to English, we evaluate our model on data in three languages, which represent different morphological types: Finnish for agglutinative, Hebrew for introflexive, and Vietnamese for isolating. Thus, we ensure that dynamic pooling is robust to different word length distributions. For English, we use \texttt{text8} (CC-BY-SA) \cite{text8}, \texttt{CC-100} (MIT) \cite{conneau-etal-2020-unsupervised} and \texttt{wiki40b} (CC-BY-SA) \cite{guo-etal-2020-wiki} as they are established benchmarks for character-level language models. For the rest of the languages, we use the corresponding subsets of \texttt{wiki40b}. To make results comparable across languages and prevent data imbalance, we limit the size of \texttt{CC-100} and \texttt{wiki40b} to the first 400M tokens of the training set and the first 2M tokens of the validation set. We retain the original splits for each dataset. 

For all datasets and languages, we follow the same pre-processing steps of \citep{text8} for creating \texttt{text8}. Specifically, for each language we keep only the characters from its script, as well as whitespace and an end-of-line. The text is lower-cased, and the digits are spelt out in the target language. For \texttt{wiki40b}, we also remove special structural markers and normalise homoglyphs. Finally, for Hebrew we also remove diacritics as they are not required to understand the text. This way, we filter out excerpts in different languages, which are known to contaminate noisy multilingual texts \citep{kreutzer2022quality}. The pre-processing scripts can be found as part of our code.

\subsection{Models}
All of our experiments, except for the scaling ablation, use 12-layer Hourglass Transformers with 2 layers in the first block, 8 layers in the second block which operates on shortened sequences, and 2 layers in the final block, following \citep{nawrot2021hierarchical}.
For every Transformer layer, the hidden dimension is $512$, the intermediate feed-forward dimension is $2048$. Self-attention is split into $8$ heads. We use a post-norm architecture, GELU activation function \cite{gelu} in feed-forward layers and the relative attention parametrisation from Transformer XL~\cite{dai2019transformer}. In total, the model has \textasciitilde41M parameters. 

The boundary predictor is a 2-layer MLP that takes a hidden state as input and outputs a scalar at every time step. For models with dynamic pooling, this module adds around 1M additional parameters. We use the SentencePiece \cite{kudo-richardson-2018-sentencepiece} library to train Unigram segmentation for every dataset separately.
We detect spikes in conditional entropy according to a window of size $k=2$, which we select from range $k\myeq 1\ldots4$ for optimal BPC on \texttt{text8}. For Gumbel Sigmoid, we set the prior probability of a boundary $\alpha$ to $0.2$ for English, Vietnamese and Hebrew, and $0.37$ for Finnish. The Gumbel temperature parameter was set to $0.5$ in all experiments. For Unigram vocabulary size, we set $|\mathcal{V}|=10000$ for English and Vietnamese and $|\mathcal{V}|=200$ for Finnish and Hebrew. We list training hyper-parameters in Appendix~\ref{app:hparams}.

\begin{table*}[t]
\centering
\resizebox{\textwidth}{!}{%
\begin{tabular}{c|rrcccc|rc|rc|rc}
\hline
 &           \multicolumn{6}{c|}{English} &
             \multicolumn{2}{c|}{Finnish}        & \multicolumn{2}{c|}{Hebrew}                             & \multicolumn{2}{c}{Vietnamese}      \\
& \multicolumn{2}{c}{text8}            & \multicolumn{2}{c}{wiki40b}          & \multicolumn{2}{c|}{cc-100} & \multicolumn{2}{c|}{wiki40b}          & \multicolumn{2}{c|}{wiki40b}          & \multicolumn{2}{c}{wiki40b} \\             
& \multicolumn{1}{c}{BPC} & \multicolumn{1}{c}{SF}
& \multicolumn{1}{c}{BPC} & \multicolumn{1}{c}{SF}
& \multicolumn{1}{c}{BPC} & \multicolumn{1}{c|}{SF}
& \multicolumn{1}{c}{BPC} & \multicolumn{1}{c|}{SF}
& \multicolumn{1}{c}{BPC} & \multicolumn{1}{c|}{SF}
& \multicolumn{1}{c}{BPC} & \multicolumn{1}{c}{SF} \\
             \hline

Vanilla & 
1.143\hspace{\starlen} & \multicolumn{1}{r|}{(1.0x)} & 
1.091\hspace{\starlen} & \multicolumn{1}{c|}{(1.0x)} & 
1.225\hspace{\starlen} & (1.0x) & 
{0.945}\hspace{\starlen} & {(1.0x)} & 
1.274\hspace{\starlen} & (1.0x) & 
1.065\hspace{\starlen}  & (1.0x) \\ \hline

Fixed (SF=2) &
1.149\hspace{\starlen} & \multicolumn{1}{r|}{(2.0x)} & 
1.084\hspace{\starlen}  & \multicolumn{1}{c|}{(2.0x)} & 
1.224\hspace{\starlen} & (2.0x) & 
0.946\hspace{\starlen} & (2.0x) & 
1.279\hspace{\starlen} & {(2.0x)} & 
1.060\hspace{\starlen}  & (2.0x) \\

Fixed (SF=3)     & 
1.155\hspace{\starlen}  & \multicolumn{1}{r|}{(3.0x)} & 
1.093\hspace{\starlen}  & \multicolumn{1}{c|}{(3.0x)} & 
1.229\hspace{\starlen}  & (3.0x) & 
0.951\hspace{\starlen} & (3.0x) & 
1.290\hspace{\starlen} & (3.0x) &
1.068\hspace{\starlen}  & (3.0x)                      \\

Fixed (SF=4) & 
1.166\hspace{\starlen}  & \multicolumn{1}{r|}{(4.0x)} & 
1.102\hspace{\starlen}  & \multicolumn{1}{c|}{(4.0x)} & 
1.240\hspace{\starlen}  & (4.0x) & 
0.961\hspace{\starlen}  & (4.0x) & 
1.304\hspace{\starlen}  & (4.0x) & 
1.087\hspace{\starlen}  & (4.0x) \\ \hline

Gumbel & 
1.136$^\star$ & \multicolumn{1}{r|}{(4.6x)} & 
1.080\hspace{\starlen} & \multicolumn{1}{c|}{(4.7x)} &
\textbf{1.212}$^\star$ & (4.6x) &
0.941\hspace{\starlen} & \multicolumn{1}{r|}{(2.6x)} & 
\multicolumn{1}{l}{1.281} & \multicolumn{1}{l|}{\textbf{(4.7x)}} & 
1.061\hspace{\starlen} & \multicolumn{1}{r}{(4.3x)} \\ 

Entropy & 
1.138$^\star$ & \multicolumn{1}{r|}{(4.1x)} &
1.083\hspace{\starlen} & \multicolumn{1}{c|}{(4.1x)} &
1.218$^\star$ & (3.8x) & 
0.949\hspace{\starlen} & \multicolumn{1}{r|}{(4.1x)} &
1.276\hspace{\starlen} & \multicolumn{1}{l|}{(3.6x)} & 
1.072\hspace{\starlen}  & \multicolumn{1}{r}{(4.2x)} \\

Unigram & 
1.134$^\star$ &  \multicolumn{1}{r|}{(5.0x)} & 
1.078$^\star$ &  \multicolumn{1}{c|}{(5.0x)} &  
\textbf{1.212}$^\star$ & (4.8x) & 
\textbf{0.937}\hspace{\starlen} & \multicolumn{1}{r|}{(2.1x)} & 
\multicolumn{1}{l}{\textbf{1.270}$^\star$} & \multicolumn{1}{l|}{(1.9x)} & 
1.058\hspace{\starlen}  & \multicolumn{1}{r}{(4.0x)} \\

Whitespaces & 
\textbf{1.133}$^\star$ & \multicolumn{1}{r|}{\textbf{(5.7x)}} & 
\textbf{1.077}$^\star$ & \multicolumn{1}{c|}{\textbf{(5.6x)}} & 
1.214$^\star$ & \textbf{(5.2x)} & 
0.955\hspace{\starlen} &  \multicolumn{1}{r|}{\textbf{(7.9x)}} & 
\multicolumn{1}{l}{1.284} &  \multicolumn{1}{l|}{(3.8x)} & 
\textbf{1.057}$^\star$ & \multicolumn{1}{r}{\textbf{(4.4x)}} \\ \hline

\end{tabular}%
}
\caption{Language modelling results on 3 English datasets and 3 other morphologically diverse languages. For each pair of method and dataset, we report test BPC ($\downarrow$) and average shortening factor (SF; $\uparrow$). We run each experiment 3 times with different random seeds. We mark with a star ($^\star$) symbol results that are statistically better than both the vanilla Transformer baseline and fixed shortening by means of a Paired Student’s t-test with $p < 0.05$. We report results based on the best hyper-parameter configuration for each language.}
\label{tab:main}
\end{table*}

\section{Results}
\begin{figure}[t!]
\centering
\includegraphics[width=\linewidth]{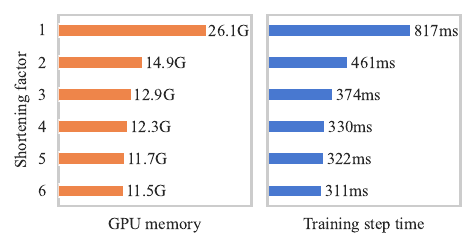}
\caption{Memory consumption and duration of a training step for different shortening factors on English \texttt{text8}. These results apply to both dynamic pooling and fixed pooling Hourglass models, as well as vanilla Transformers (for SF=1).}
\label{fig:mem}
\end{figure}

The results for the experiments on character-level language modelling are shown in \cref{tab:main}. In addition to the four proposed segmentation methods, we include a vanilla Transformer and fixed-size pooling Transformers with multiple shortening factors as baselines. Every model is evaluated with respect to two metrics: bits per character (BPC; $\downarrow$) and shortening factor (SF; $\uparrow$). The former measures the negative log-probability of the language model predictions, and thus its quality; the latter measures the average reduction of the sequence length in intermediate layers, and thus the model efficiency.
Figure~\ref{fig:mem} shows how higher SF translates to lower training time and memory consumption in practice, as measured on a common GPU with an optimised model implementation.

\paragraph{Segmentation Methods}
In all the English evaluation benchmarks (\texttt{text8}, \texttt{wiki40b}, and \texttt{CC-100}), both whitespace-based and Unigram-based segmentations achieve the lowest BPC, outperforming both vanilla and fixed-pooling Transformers by statistically significant margins.\footnote{We indicate with a $\star$ wherever this is the case according to a Paired Student's t-test with $p < 0.05$.} Moreover, the same two methods achieve the highest degrees of shortening. Note that for equivalent SFs, fixed-size pooling becomes detrimental to performance. 
The approaches based on entropy spikes and Gumbel-Sigmoid are generally inferior to the alternatives for dynamic pooling.
However, for comparable shortening factors, they always outperform vanilla and fixed-pooling Hourglass models.
Moreover, they make the fewest assumptions about the data and the availability of external supervision, so they might be appropriate for other domains (such as speech and vision) in future work. In general, providing a Transformer with the correct inductive bias for pooling variable-size segments not only facilitates scaling but also enhances prediction quality.

Notably, the gains resulting from whitespace segmentation are not identical in all languages, due to their inherent differences in morphological types and average word length. Shortening Factors for this method range from 3.8$\times$ in introflexive Hebrew, to 7.9$\times$ in agglutinative Finnish, whereas isolating Vietnamese and mildly fusional English lie in between with 4.4$\times$ and 5.7$\times$, respectively. The larger SFs of dynamic pooling methods translate into higher training speed, from 1.7$\times$ for Unigram in Hebrew to over 2.5$\times$ for whitespaces in English, while simultaneously lowering BPC. Overall, the gains from dynamic pooling are robust cross-lingually, but the optimal segmentation method may vary.

\begin{figure}[t!]
\centering
    \resizebox{1.0\linewidth}{!}{\includegraphics{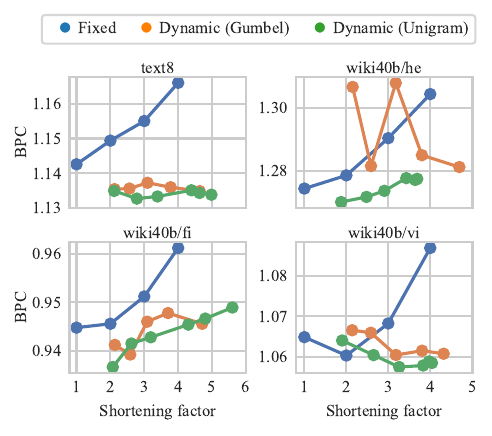}}
\centering
\caption{Test BPC ($\downarrow$) and shortening factor (SF; $\uparrow$). The higher the SF, the more efficient the model is (cf. Figure~\ref{fig:mem} in the Appendix). SF increases with higher vocabulary size (Unigram) or smaller prior boundary probability (Gumbel). Dynamic pooling methods shift the Pareto front, i.e., increase performance for the same efficiency (and vice versa). Note that fixed-pooling at $k\myeq 1$ corresponds to the vanilla Transformer model.}
\label{fig:unigram}
\end{figure}

\paragraph{Efficiency--Performance Pareto Front}
While both low BPC and high SF are desirable, there exists a trade-off between them which is specific to each boundary prediction method. Hence, the ideal model should strike the right balance to improve in both respects simultaneously. Intuitively, vocabulary size in Unigram and the prior $\alpha$ in Gumbel-Sigmoid provide easily controllable knobs to study this interaction: as they change, so does the shortening factor. In \cref{fig:unigram}, we plot BPC and SF for six vocabulary sizes (200, 500, 1k, 3k, 5k, 10k) and five $\alpha$ values (0.20, 0.25, 0.30, 0.37, 0.45) and compare them with fixed-size pooling in Hourglass Transformers. Manifestly, dynamic pooling enhances the Pareto front by finding more optimal trade-offs between efficiency and performance. Moreover, while fixed pooling follows a similar trend cross-lingually, dynamic pooling behaves more idiosyncratically: e.g. BPC in Vietnamese and English surprisingly improves with higher SFs. During our study of the Efficiency--Performance Pareto Front, we noticed that the Gumbel-Sigmoid pooling approach exhibits greater instability compared to the Unigram-based pooling method. This can be observed through artifacts such as the spikes in BPC for Hebrew, depicted in \cref{fig:unigram}.

\paragraph{Time and Space Complexity}
To capture the concrete gains in efficiency of models with higher SFs, we have measured the memory consumption and training time of our PyTorch implementation of \texttt{text8} models on a typical GPU (NVIDIA GV100 32GB). The results in \cref{fig:mem} apply to dynamic-pooling (Gumbel, Whitespace, Unigram, and Entropy), fixed-pooling, and vanilla Transformers (only for SF=1). Note that these results are identical for both fixed-pooling and dynamic-pooling Hourglass for the same SF as the cost of the boundary predictor is negligible. With a shortening factor $SF=2$, the model reduces both memory consumption and training time by over $40\%$, compared to a vanilla Transformer. At $SF=4$, where dynamic-pooling Hourglass still achieves superior BPC scores, resource consumption is reduced between $50\%$ and $60\%$ and training is 2.5$\times$ faster. This allows models to increase in size with the same compute budget (which depends on the hardware), while simultaneously benefiting their performance.

\begin{figure}[t]
\centering
    \includegraphics{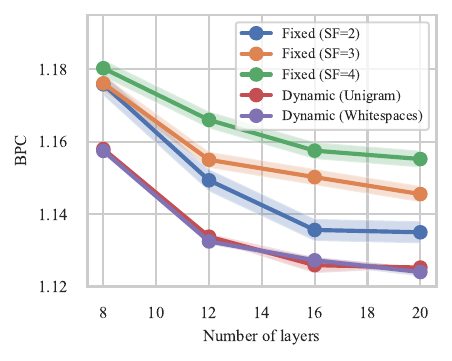}
\centering

\caption{Test BPC on \texttt{text8} plotted against the number of Transformer layers  for different shortening methods. We use two layers in the first and last transformer block and only scale the middle, downsampled block. There are 28M parameters in models with 8 layers, up to 69M parameters in models with 20 layers. For all variants we observe performance gains with dynamic pooling.}
\label{fig:scaling}
\end{figure}

\paragraph{Scaling the Model}
We investigate if dynamic-pooling Transformers scale well in terms of model size, by adding more layers in the middle block (\cref{fig:scaling}). We focus on this block as it increases the model depth (and hence its capacity) while retaining a higher efficiency due to operating on shortened sequences. We find that the gains from dynamic pooling are consistent across all numbers of layers. Extrapolating from the trends, dynamic pooling holds promise to continue providing benefits even in extremely large language models. 

\paragraph{Average-pooling vs Sub-sampling}
As an ablation, we also compare two different methods to represent groups of tokens when shortening the input sequence length: average pooling, used in our experiments, and sub-sampling, i.e. selecting only the last token as a representative for each group. As it emerges from \cref{tab:shortening}, average pooling yields superior performance in all models, including both fixed and dynamic pooling Transformers.

\begin{table}[ht!]
\centering
\begin{tabular}{lcc}
\hline
                    & \multicolumn{2}{c}{Shortening} \\ \cline{2-3} 
Segmentation & Avg-Pooling            & Sub-sampling           \\ \hline
Fixed (SF = 2)      & \textbf{1.149}        & 1.180                     \\
Entropy             & \textbf{1.138}        & 1.151                     \\
Whitespaces         & \textbf{1.133}        & 1.144                     \\ \hline
\end{tabular}
\caption{BPC results on \texttt{text8} for two shortening methods (average-pooling and sub-sampling) and three segmentation methods.}
\label{tab:shortening}
\end{table}

\paragraph{Other Efficient Transformer Models}
Finally, we remark that our method differs from most efficient Transformer algorithms, which reduce the quadratic complexity of attention \cite{child2019generating,leethorp2022fnet,choromanski2021rethinking,wang2020linformer}, as it focuses on length reduction. While previous efficient variants tend to trade quality for efficiency, we have shown that the dynamic-pooling mechanism improves both simultaneously in our experiments. Moreover, \citet{nawrot2021hierarchical} has shown that combining both strategies yields further gains.

\section{Related Work}

\paragraph{Dynamic RNNs}
Our approach is inspired by variants of RNNs that process sequences at varying time scales by introducing a hierarchy of hidden units. For instance, RNNs that mimic speed-reading by introducing hidden units that can skip over some input elements \citep{skiprnn, skimrnn}. Similarly, \citep{hierarchicalrnn} discovers the latent hierarchy of an input sequence using a stack of LSTMs. Each layer is equipped with a binary gate responsible for hard boundary detection, where lower-level boundaries determine state updates made by higher-level layers. Whenever the detector ends a segment, its representation is fed to the upper layer.

Early slow- and fast-changing units were already described by \citep{hihi1995hierarchical}. Similarly, Clockwork RNN \citep{koutnik2014clockwork} introduces a hierarchy of hidden state units that make transitions at a set of different, fixed frequencies. Adaptive Computation Time networks perform a different amount of computation on each sequence item \citep{graves2016adaptive}. Both ideas were combined in Fast-Slow RNNs \citep{mujika2017fast} which can choose a heavy or light transition between timesteps. 

\paragraph{Pooling Transformer models}
While pooling blocks in Transformers are related to slowly varying units in RNNs,
their operation is different.
RNNs suffer from unreliable transport of information across long time spans.
Units that act like skip-connections over time can help them to carry information~\citep{krueger2017zoneout}.
In a Transformer network, a unit at time $t$ can directly communicate with any other unit, including previous ones,
and we find it important to confirm the benefits of dynamic pooling in Transformer models.

Perhaps the most similar approach to ours is Funnel Transformer~\citep{dai2020funnel} which uses a similar, hourglass-shaped Transformer architecture. After passing through the first block, the data is pooled at a fixed rate, processed by the deep middle Transformer block, and up-sampled for the last block. Canine~\citep{clark2021canine} has a similar three-part architecture, and processes Unicode inputs, which are downsampled with Transformer and convolution layers. \citep{tay2021charformer} implements gradient-based subword tokenization within a Transformer model,
which learns dynamic groupings of tokens into fixed-size groups. In \citep{bai2021segatron}, sentence and paragraph boundaries were used as additional conditioning for the model.

\paragraph{Boundary Detection}
We investigate boundaries provided by an external model, derived directly from the data, or top-down from the model's entropy.
\citep{kreuk2020selfsupervised} shows a bottom-up approach to phoneme segmentation task combining contrastive learning \citep{oord2019representation} with a method for boundary detection based on dissimilarity between subsequent frames. It was later extended by \citep{bhati2021segmental} to segment the sequence of speech frames dynamically. Recently, \citep{cuervo2022variable} introduced a hierarchical sequence processing model in which units in the upper layer operate on a dynamically shortened sequence, with the shortening guided by a boundary prediction model.

\citep{rocki2016surprisal} control the activity of LSTM gates with the model's output cross-entropy. \citep{alpay2019preserving} used a similar mechanism based on information content to guide the copying of individual activations in an LSTM network. Similarly, we employ the entropy of model predictions to choose where to insert boundaries.

\section{Conclusions}
We proposed a new family of language models that pool variable-size segments of tokens in the intermediate layers in order to enhance the efficiency and performance of the Transformer architecture. In particular, we learn a boundary predictor either end-to-end through stochastic re-parameterisation, through supervision (obtained from subword tokenization or spikes in the conditional entropy), or based on linguistic boundaries such as words. We evaluate this model extensively on multiple language modelling benchmarks in English and in other typologically diverse languages: Finnish, Hebrew, and Vietnamese. Compared to vanilla Transformers and fixed pooling, we observe a significant decrease in model perplexity as well as time and space complexity. This opens up the perspective to develop Transformer models capable of computing language both hierarchically, with the same abstractions humans perform at different levels of linguistic structure, and conditionally on the information content of each segment.

In the future, our dynamic-pooling Transformer can be combined with methods relying on external memory \citep{wu2022memorizing}, encoders operating at a fine resolution \citep{xue2022byt5,tay2021charformer}, and more generally any task with long-context inputs \cite{shaham2022scrolls}. This may further facilitate the scalability of current language modelling architectures. 

\section{Limitations}
\paragraph{Linguistic variation}
Our results are highly dependent on the target language and its morphology. For example, word boundaries might seem like an obvious choice for dynamic segmentation, and in fact they achieve the best performance in English and Vietnamese. However, for some languages like agglutinative Finnish, whitespaces are less frequent, which is detrimental to model performance. Explicit word boundaries are not available for all scripts. For example, in Chinese characters, or in modalities other than text like speech or vision, there is no obvious equivalent to whitespaces. However, segmentation based on stochastic re-parameterisation, subword tokenizers and spikes in conditional entropy overcomes these limitations.

\paragraph{Contiguous segments}
In its current formulation, dynamic pooling only allows for merging contiguous segments of tokens in a sequence. However, this is not ideal for morphology types like Hebrew where morphemes are discontinuous: vowels are interspersed between consonant roots for inflection. Moreover, future works should consider higher levels of linguistic structure than words, such as dependency trees, for pooling. In this case, discontinuous segments may be necessary to handle non-projective syntactic dependencies.

\paragraph{Independent boundary decisions}
The decision to emit a boundary at time step $t$ depends on previous boundaries only indirectly through the hidden representation of the first Transformer block, as this preserves the efficiency of the boundary predictor. Instead, a recurrent model could be explicitly conditioned on previous boundary decisions, which however would negatively affect the time complexity of the language model.

\section*{Work contribution of authors}
The idea of training the models with pooling of variable-length segments was discussed among the authors while Jan Chorowski was at the University of Wrocław. Experiments were performed by Piotr Nawrot while he was employed in a research grant at the University of Wrocław, under the supervision of Adrian Łańcucki and Edoardo M.\ Ponti. The manuscript was written by Piotr Nawrot, Adrian Łańcucki and Edoardo M.\ Ponti.

\section*{Acknowledgements}
This work was supported in part by the UKRI Centre for Doctoral Training in Natural Language Processing, funded by the UKRI (grant EP/S022481/1) and the University of Edinburgh, School of Informatics and School of Philosophy, Psychology \& Language Sciences; and the Polish National Science Center under the OPUS-18 2019/35/B/ST6/04379 grant.

\bibliography{custom}
\bibliographystyle{acl_natbib}

\clearpage
\newpage
\onecolumn

\appendix
\section*{Appendix}

\section{Frequently Asked Questions}
\label{app:faq}

\subsection{Pros and Cons of shortening methods}

\begin{table}[ht!]
\centering
\resizebox{\columnwidth}{!}{%
\begin{tabular}{l|l|l}
\hline
            & \multicolumn{1}{c|}{Pros}                                                                                                                               & \multicolumn{1}{c}{Cons}                                                                                                                                                   \\ \hline
Fixed       & - Simple                                                                                                                                                & - Sub-optimal results, especially for SF > 2                                                                                                                               \\ \hline
Whitespaces & \begin{tabular}[c]{@{}l@{}}- Linguistically inspired\\ - Does not require a boundary predictor\end{tabular}                                             & \begin{tabular}[c]{@{}l@{}}- Not available in all languages, e.g., Chinese\\ - No control over SF\end{tabular}                                                             \\ \hline
Entropy     & \begin{tabular}[c]{@{}l@{}}- Better performance than Fixed\\ - Suitable for other modalities such as speech and vision\end{tabular}                     & \begin{tabular}[c]{@{}l@{}}- Requires a boundary predictor\\ - Worse than Unigram and Gumbel\end{tabular}                                                                  \\ \hline
Unigram     & \begin{tabular}[c]{@{}l@{}}- Best trade-off between efficiency and performance\\ - Shown to align well with morphological units\end{tabular}            & \begin{tabular}[c]{@{}l@{}}- Requires a boundary predictor\\ - Works only in sequential discrete data\\ - Requires training a tokenizer up-front\end{tabular} \\ \hline
Gumbel      & \begin{tabular}[c]{@{}l@{}}- Good trade-off between efficiency and performance\\ - Suitable for other modalities such as speech and vision\end{tabular} & \begin{tabular}[c]{@{}l@{}}- Requires a boundary predictor\\ - High variance performance\end{tabular}                                                                                  \\ \hline
\end{tabular}%
}
\caption{Pros and cons of different shortening methods. SF is a shorthand for Shortening Factor.}
\label{tab:summary}
\end{table}

\subsection{What is the ultimate segmentation method?}
While Whitespace offers the best performance in many cases, this is not always true even in the linguistic domain. In agglutinative languages (e.g., Finnish), words are longer than in English, which has a detrimental effect on the Whitespace method. For such languages, other dynamic methods that allow for controlling the shortening factor (SF), such as Unigram, are better suited. Moreover, languages with non-Latin scripts (like Chinese) may lack explicit whitespaces. For modalities different from text, such as speech and vision, Gumbel and Entropy are to be favoured as they do not assume the discreteness of the input sequence.

\subsection{Why evaluating on language modelling rather than downstream tasks?}
Since we present a proof of concept for dynamic-pooling Transformers, we limit the experiments to language modelling because: 1) it is a foundational NLP task; 2) previous efficient Transformer variants were evaluated on similar benchmarks. Crucially, there is a strong correlation between performance in language modelling and downstream tasks.

\subsection{How do you ensure that the results are reliable?}
Our code is based on the optimised, open-source implementation of Transformer-XL from NVIDIA (Apache 2.0 License), which reproduces the scores reported by \cite{dai2019transformer}. Our implementation of the fixed-pooling Hourglass Transformer model similarly reproduces the results from \cite{nawrot2021hierarchical}. We make our code publicly available, under the Apache 2.0 License, inheriting from the original source, to ensure the reproducibility of our results. Moreover, memory utilisation was measured by controlling resource allocation on GPUs (\cref{fig:mem}) rather than through a naive \texttt{nvidia-smi} readout, as this would overestimate the reserved buffers.

\section{Hyper-parameters}
\label{app:hparams}
Following \citep{dai2019transformer}, we train for $2 \cdot 10^5$ steps with a batch size of 8 and a learning rate $2.5 \cdot 10^{-4}$ on 2x NVIDIA RTX 3080. Each training run took from approximately 12h to 30h, depending on the configuration. We use a linear warm-up schedule for the first 4k steps, followed by a single-cycle cosine scheduler. We use an Adam optimiser with $\beta_1 = 0.9$, $\beta_2 = 0.999$ and $\epsilon = \expnumber{1}{-8}$, and clip the gradients at $0.25$. We apply a $0.1$ dropout rate in the attention matrix and feed-forward layers. Before every epoch, we cyclically shift the text stream, divide it into non-overlapping chunks of 2048, and shuffle. During the evaluation, to provide context to the model, we split the test set into partially overlapping sequences of size $l = 2048$ with a step size of $512$ and calculate the model perplexity only over the last $512$ tokens.

\end{document}